\documentclass[runningheads]{llncs}
\usepackage{graphicx}
\usepackage{amsmath,amssymb} % define this before the line numbering.
\usepackage{color}
\usepackage[width=122mm,left=12mm,paperwidth=146mm,height=193mm,top=12mm,paperheight=217mm]{geometry}
\begin{document}
% \linenumbers
\pagestyle{headings}
\mainmatter
\def\ECCV16SubNumber{***}  % Insert your submission number here

\title{MathDivide: Improved mathematical reasoning by large language models} % Replace with your title

\titlerunning{\textit{MathDivide: Improved mathematical reasoning by large language models}}

\authorrunning{\textit{Srivastava et al.}}

\author{Saksham Sahai Srivastava \and Ashutosh Gandhi}
%\institute{}

\institute{Department of Computer Science\\
University of Colorado Boulder\\
\email{\{saksham.srivastava, ashutosh.gandhi\}@colorado.edu}}

\maketitle

\begin{abstract}
Large language models have been proven to be capable of handling complex linguistic and cognitive tasks. Therefore their usage has been extended to tasks requiring logical reasoning ability such as Mathematics. In this paper, we propose a prompting technique called MathDivide that breaks down the mathematical problem into simpler subproblems. Each of the subproblems is formulated as an algebraic expression whose value is evaluated by the Python code generated by the LLM for the corresponding algebraic expression. The values fed to the Python code are the numerical values provided in the problem statement. The solutions for the subproblems are composed together to obtain the final answer for the problem statement. Finally, the final answer is compared to the correct answer. If the final answer matches the correct answer, it is produced as output else a refinement prompt is fed to the LLM. We experiment with this prompting technique on both closed-source LLM models and open-source LLM models using GSM8K dataset. The results obtained demonstrate that MathDivide was able to significantly outperform the leading prompting technique called Mathprompter.


\end{abstract}

\section{Introduction}
Recent years have been witness to significant advancement in the field of natural language processing. The groundbreaking work of "Attention is all you need"\cite{vaswani2017attention} which utilized the idea of self-attention was the foundation for large language models. This progression in the field of NLP was focused on how large amounts of text data can be used to train algorithms such that they become capable of understanding and generating human-like text. The complex architecture and vast amount of training data empower the LLMs to perform linguistic and cognitive tasks requiring computational intelligence. However, recent research shows that LLMs are also proficient in performing tasks that require analytical and logical reasoning\cite{pan2023logic,chen2023learning}. Solving mathematics problems is one such task that requires analytical and logical reasoning ability. 

\par In our work we try to improve upon the mathematical reasoning capability of the pre-trained large language models. The pre-trained LLMs can be both open-source\cite{touvron2023llama,touvron2023llama2,MosaicML2023Introducing,penedo2023refinedweb,gpt-j,roziere2023code} as well as closed source\cite{openai_gpt35_turbo,openai_gpt4,chowdhery2023palm,anil2023palm,lewkowycz2022solving,bai2022constitutional,chen2021evaluating}. It has been noticed from previous literature that LLMs are good at in-context learning\cite{dong2022survey} which led to the evolution of different prompting techniques such as zero-shot\cite{kojima2022large}, few-shot\cite{brown2020language}, chain-of-thought(CoT), etc. Kojima et al.\cite{kojima2022large} showed that the zero-shot-CoT prompting approach significantly improves the reasoning capability of LLM, therefore in this paper, we develop a zero-shot-CoT-based prompting technique. We will extend the idea of how a student when given a mathematical problem solves it in a step-by-step process to LLMs. At each step, the student intends to solve a subproblem whose solution is required to move to the next step and ultimately arrive at the final answer for the original problem. We take this approach because the hidden potential in solving the problem by breaking it into smaller subproblems was demonstrated by Shridhar et al.\cite{shridhar2022distilling} where they showed that this approach can even aid smaller models such as GPT-2 to surpass the performance of larger models such as GPT-3 6B for various reasoning datasets. 

\par The mathematical problem is often formulated as an algebraic expression which is deemed to be compact and more representative of the problem details. Not only that, Imani et al.\cite{imani2023mathprompter} showed that crafting the mathematical word problem as an algebraic equation can even beat the performance of state-of-the-art zero-shot-CoT approach\cite{kojima2022large} for mathematical reasoning tasks. We combine the strength of the two approaches to design a prompt in such a way that the LLM splits the problem into subproblems that can be solved in sequential steps and for each sub-problem an algebraic expression is constructed whose numerical value needs to be determined(using the numerical values provided in the original problem). We feed the problem statement as well as this prompt to the LLM. In this approach, the LLM is also leveraging the power of the chain of thoughts indirectly.

\par %Some open-source LLMs which we use a benchmark may not have the capability of performing mathematical operations, therefore 
We add an instruction to our prompt which directs the LLM to produce a Python code snippet for the subproblem and perform the mathematical operation using the Python code. Finally, the prompt instructs the LLM to compose the solution of all the sub-problems and produce the final answer for the original problem. If the final answer does not match the gold correct answer then we provide a refinement prompt to the LLM highlighting which subproblem(s) was incorrectly solved. This human-based feedback has a great potential to drive the LLM towards the right answer.
\par In our approach, we try to make the LLM mimic the human problem-solving strategy when solving a mathematical problem by innovatively combining a chain-of-thoughts approach with algebraic expression formulation, breaking down complex problems into simple sub-problems, making use of python-code snippets to ensure computational precision and making use of human feedback based refinement. The strength of human-based feedback lies in helping the LLM to learn in an in-context fashion which adapts it to avoid making the same mistake for future problems. This nice blend of structured problem decomposition, precise computation, and iterative refinement positions makes our solution better as compared to existing methodologies in LLM-based mathematical problem-solving.

\section{Related Work}
Noorbakhsh et al.\cite{noorbakhsh2021pretrained} presented a technique where they leveraged pre-trained LLMs which were originally trained on language tasks, for solving mathematical symbolic problems like differentiation and integration. They fine-tuned the LLM for similar mathematical tasks using a small training set size. This resembled the knowledge transfer capability of the LLMs from natural language processing to mathematical reasoning. Later, Yuan et al.\cite{yuan2023scaling} studied the impact of factors such as pre-training loss, amount of supervised data, and amount of augmented data on the mathematical reasoning capability of LLMs and their exploration revealed that lowering the pre-training loss can help LLM to solve the math problem accurately. Yamauchi et el.\cite{yamauchi2023lpml} proposed an LLM-prompting markup language (LPML) which is a structured language like XML that facilitates combining of chain-of-thoughts approach with an external computation tool(Python REPL). This way a structured output was produced from the LLM which was effective in guiding the LLM to rectify any reasoning or calculation mistakes. A novel idea of tutoring the LLM was also used by Liu et al.\cite{liu2023novice} to improve its performance for mathematical reasoning. They assessed the LLM's conceptual understanding of mathematical problems both as a learner and a tutor, however, they noticed that LLMs were not great at identifying correct misconceptions. This suggested a gap in LLM's proficiency in mimicking a human-like behavior of learning and tutoring. In our approach we take advantage of the strength lying in structured output as shown by \cite{yamauchi2023lpml} and therefore we direct the LLM to produce algebraic expressions for each subproblem like a grade-school mathematics student would do. This also enhances the LLM's ability to behave as a learner and identify misconceptions.
\par In the realm of refinement of the output produced by LLM, Madaan et el.\cite{madaan2024self} developed a self-refine technique where the proposed algorithm would enable the LLM to enhance its outputs through self-generated feedback iteratively. This approach is completely autonomous and utilizes only the model's initial outputs as a basis for continuous improvement and does not require any additional supervised data or training. But for prompt refinement, Zhang et al.\cite{zhang2024prefer} proposed a technique called \textit{PREFER} where they employed a cycle of feedback and refinement in such a way that the LLM was able to leverage its capabilities to generate improved prompts. They took advantage of the process of prompt bagging in their technique which incorporated both forward and backward thinking for output evaluation. In an attempt to generate effective prompts, Billa\cite{billa2024supervisory} proposed a dual LLM system where one of the LLM is a 'generator' and the other is a 'corrector'. The 'generator' LLM would perform the task and the 'corrector' LLM would provide feedback and generate improved prompts, thus in a way collaborating over time to generate prompts that guide the LLM towards the correct answer. Our approach of using a human-based feedback mechanism can introduce more nuanced insights to the LLM, helping them in a way to identify and correct the errors. It is highly probable that the automated refinement techniques might not be able to recognize complex reasoning errors or have contextual misunderstanding while our approach allows for more accurate, context-aware, and user-centric refinements. %Also, modern LLMs have the capability of in-context learning, thus after the manual refinement the LLM is not expected to make the same error for future questions in the same chat window.
%\colorbox{red}{This text is inside a red box.}

\section{Methods}
Our methodology starts with decomposition of a mathematical problem($\mathcal{M}$) into smaller, simpler and manageable sub-problems($\mathcal{M}_1,\mathcal{M}_2,\dots,\mathcal{M}_k$). We parse the natural language representation of the original problem to the LLM so that it gets to identify the key components in the problem and their relationship with each other. A prompt($p$) is also appended at the end of the problem statement. Each component of the problem is considered as a subproblem($\mathcal{M}_i$) to be solved. Based on the prompt $p$, the LLM formulates an algebraic expression($e_i$) for each subproblem($\mathcal{M}_i$) such that it encapsulates the core mathematical operation required to solve the task ($\mathcal{M}_i$). After the construction of $e_i$, the LLM generates a Python code($p^c_i$) such that it computes the numerical value of $e_i$ when each of the variables in $e_i$ is substituted with their values. This means that each variable in the expression $e_i$ will be a parameter for the Python function $p^c_i$. If the subproblems are related to each other then each subproblem is solved in sequential steps which interprets to first $\mathcal{M}_1$ being solved, then $\mathcal{M}_2$ is solved and finally, $\mathcal{M}_k$ is solved. If the computation of subproblems does not depend on the evaluation of other subproblems then each subproblem is computed independently. The final answer $s$ to the original problem is obtained by composing the answer obtained from each subproblem($s_i$).

\par The next step is to compare the value of $s$ with the actual desired output $z$. If $s\neq z$, then we feed the LLM with a refinement prompt($p_r$) which is based on human feedback and exactly highlights the component where the LLM went wrong. This also helps the LLM to have an in-context learning. If $s=z$, no such refinement prompt is needed. Figure \ref{fig:fig1} shows the architecture diagram of our technique. The prompt $p$ used in our technique is given below.\newline

\begin{figure}
    \centering
    \includegraphics[width=1\textwidth]{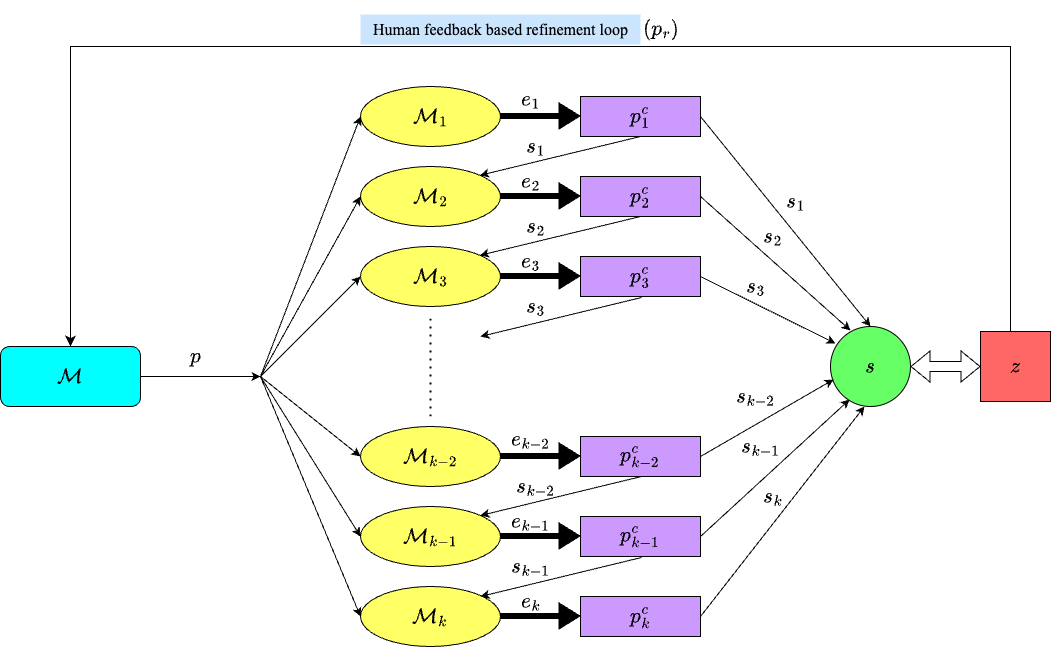}
    \caption{Architecture Diagram of MathDivide Prompting technique}
    \label{fig:fig1}
\end{figure} 

\noindent\textbf{Prompt $p$:}

\begin{quote}
\textit{Given the mathematical problem: $\mathcal{M}$, your task is to solve it step by step. Start by breaking down the problem into smaller, manageable sub-problems. For each sub-problem identified:}

\begin{enumerate}
    \item \textit{Describe the sub-problem in your own words and identify the key components and their relationships.}
    \item \textit{Formulate an algebraic expression that captures the core mathematical operation needed to solve this sub-problem. Denote this expression as $e_i$.}
    \item \textit{Generate Python code that computes the numerical value of $e_i$. Assume each variable in $e_i$ as input parameters to your Python function.} 
    \begin{verbatim}def compute_ei(var1, var2, ...): 
        # Your Python code here
        return result\end{verbatim}
    \item \textit{Execute the Python code to solve the sub-problem and provide the solution.}
\end{enumerate}

\textit{Proceed sequentially if the sub-problems depend on each other. If they are independent, you may solve them in any order. Once all sub-problems are solved, combine their solutions to provide the final answer to the original problem $\mathcal{M}$.}

\textit{If your final answer does not match the expected solution, you will receive specific feedback indicating where adjustments are needed. Use this feedback to refine your approach and improve your solution.}
\end{quote}

\par The prompt $p$ is a bit long but since the LLMs that we would be considering for our experimentation have a context window of more than 2000 tokens it will not lead to an issue where the LLM is only focusing on the later part of the prompt. 

\section{Experiment}
We compare the performance of our prompting technique i.e. MathDivide with Mathprompter\cite{imani2023mathprompter} technique which previously was able to successfully beat the performance of the state-of-the-art zero-shot-CoT\cite{kojima2022large} approach. We evaluate the performance of our prompting technique with both proprietary models - i) GPT-3.5-turbo\cite{openai_gpt35_turbo} ii) GPT-4\cite{openai_gpt4} as well as open-source LLM models - i) Llama2\cite{touvron2023llama2} - 7B parameters ii) Llama3\cite{llama3} - 8B parameters. 

\par We utilize ChatGPT for harnessing the power of both GPT-3.5-turbo and GPT-4 LLM models. Since the GPT models do not provide cheap API access as well as our technique requires manual human-based feedback, we conducted the experimentation with only the first 250 math word problems in  GSM8K\cite{openai_grade_school_math} dataset. Although, the Mathprompter technique was evaluated for MultiArith dataset\cite{roy2016solving} which is a subset of MATH dataset\cite{hendrycks2021measuring} and used 175B parameter LLM for few-shot and few-shot-CoT settings and 540B parameter LLM(PaLm 540B) for zero-shot and zero-shot-CoT settings but in order to have a fair and direct comparison we evaluate the performance of Mathprompter\cite{imani2023mathprompter} technique on the same LLM models and dataset. 

\par To run the Llama2 and Llama3 models we utilized an open-source project called Ollama\cite{ollama}. Ollama\cite{ollama} provides a memory-quantized version of many open-sourced LLMs through an API. For our project, we used llama2:7b-chat-q2\_K and llama3:8b-instruct-q2\_K versions of Llama2 and Llama3 respectively. We implemented a simple Python script to parse the GSM8K dataset and extract and question and answer. We append our custom prompt to each question and call Ollama\cite{ollama} API to get the LLM response. The response is then stored and compared to the answer to verify if LLM was able to solve it or not. In case of the MathDivide method for each incorrect response, we provide specific feedback prompts to correct the response and keep the count of the number of refinement prompts required to get the correct response. Since, ChatGPT-3.5 and Llama models do not have the ability to execute python codes, we manually executed the code snippets generated by these LLMs. We fed the numerical answer obtained after code execution back to the LLMs. 

\par We use accuracy as the evaluation metric as it serves as an essential baseline for estimating the model's effectiveness in producing correct answers. We compute accuracy for up to 3 human-based feedback loops. In other words, if the LLM can output the correct answer for a problem within 3 refinement prompts we consider the problem to be correctly solved by our technique. The mathematical expression for the accuracy of our technique can be given as:
\[
    Accuracy\% = \frac{\text{No. of problems solved within 3 refinement prompts}}{\text{Total no. of problems fed to the LLM}}\times 100\%
\]

\begin{figure}[ht]
    \centering
    \begin{minipage}{0.49\textwidth}
        \centering
        \includegraphics[width=\textwidth]{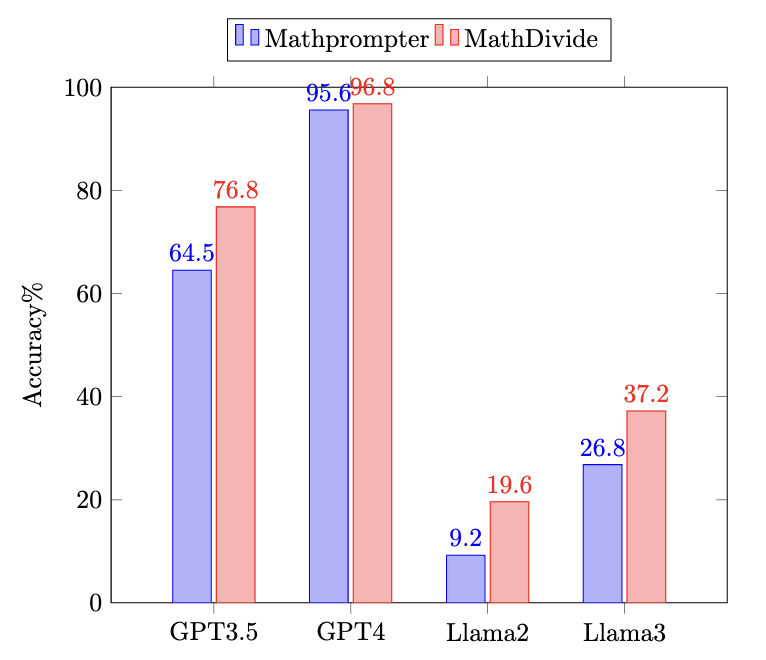}
        \caption{Accuracy Comparison of MathDivide with Mathprompter}
        \label{fig:accuracy1}
    \end{minipage}\hfill
    \begin{minipage}{0.49\textwidth}
        \centering
        \includegraphics[width=\textwidth]{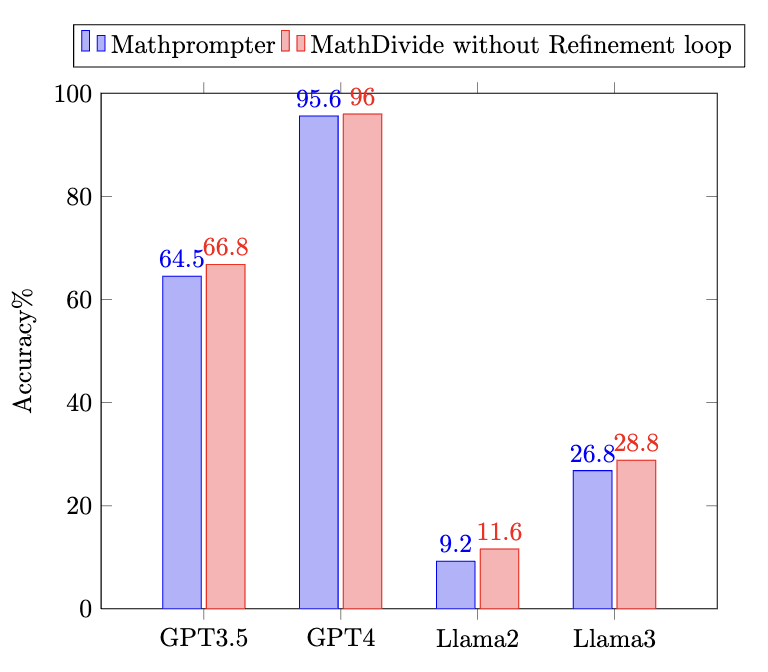}
        \caption{Accuracy Comparison of MathDivide without refinement loop with Mathprompter}
        \label{fig:accuracy2}
    \end{minipage}
\end{figure}

For the mathprompter technique, the accuracy is evaluated without any refinement loop. The accuracy values obtained for both prompting techniques on the chosen LLM models and dataset is depicted in Figure-\ref{fig:accuracy1}. We also evaluate the performance of the MathDivide technique when used without the refinement loop as shown in Figure-\ref{fig:accuracy2}. The results indicate that the proposed prompting technique MathDivide was able to beat the performance of Mathpromter technique.

\par We observe that both the proprietary models ChatGPT-3.5 and ChatGPT-4 demonstrated a significantly higher accuracy for both the prompting techniques than the Llama2 and Llama3 models. This signifies that the OpenAI's GPT models are better than Meta's Llama models at handling tasks involving mathematical reasoning. The exact reason for this is unknown due to the unavailability of architecture and training data information of OpenAI's GPT models. Since, MathDivide technique which incorporated a refinement loop showed better accuracy than the Mathprompter technique for all the four LLM models, it can be deduced that the refinement loops are considerably effective in enhancing the performance of large language models for tasks involving analytical and logical reasoning.

\par Figure-\ref{fig:accuracy2} shows that even without the refinement loop, MathDivide technique was able to perform better as compared to Mathprompter. This indicates that the fundamental approach of breaking the complex problem into simpler sub-problems is a beneficial approach for Mathematical reasoning tasks. The problems solved correctly after using refinement loops generally required a single refinement prompt: "\textit{Check the calculations}". 

\par The experiment was conducted on small dataset of size 250 problems from GSM8K dataset\cite{openai_grade_school_math}, therefore testing on a larger and diverse dataset(including real-world complicated math word problems) could provide more comprehensive insights into the robustness and generalizability of these prompting techniques. Furthermore, the use of human feedback-based refinement and the inability of some LLM models to execute programs forces the act of manually feeding prompts to the LLM. This presents a challenge to conducting experimentation on huge amounts of data. Therefore, the exploration of automated refinement techniques for Math word problem-solving tasks would be immensely beneficial for making the experimentation process more scalable. 

\par This research can also be extended to studying the real-time learning and adaptation of large language models. In real-time learning, the LLM learns from each human interaction and adapts to enhance its performance and utility without requiring any sort of retraining. This could involve developing techniques that can aid the LLMs to incrementally learn from the refinement prompts and immediately apply the learned strategy to the new problem given to them.   

\section{Conclusion}
This paper proposes a novel prompting technique called MathDivide which significantly improved the mathematical reasoning capability of large language models. This technique solves a complex mathematical problem by breaking it down into smaller and simpler sub-problems. It also leverages a human-feedback-based refinement loop to further enhance its accuracy. This technique was able to beat the performance of a leading prompting technique called Mathprompter\cite{imani2023mathprompter} which had previously exhibited a better accuracy compared to the state-of-the-art\cite{kojima2022large} zero-shot-Cot prompting approach. Therefore, the proposed technique highlights the significance of solving the math problem in a structured manner.

\par Lastly, it is important to address the ethical implications of our research. We ensure fairness by comparing the proposed prompting technique with state-of-the-art math prompting technique on the same sets of LLMs and dataset. We maintain transparency of this research work by providing clear and detailed descriptions of the methods and dataset set used for experimentation. This allows for replication and verification of our proposed work by other researchers in the field. This also facilitates accountability in our processes. Lastly, our research does not have any adverse social implications. In this manner, we aim to uphold the highest ethical standards by ensuring that our research contribution to LLMs is both innovative and responsible.
\bibliographystyle{splncs}
\bibliography{egbib}
\end{document}